\documentclass{article}

\usepackage{microtype}
\usepackage{graphicx}
\usepackage{subfigure}
\usepackage{booktabs}
\usepackage{amsmath}
\usepackage{amsfonts}
\usepackage{amsthm}
\usepackage{dsfont}
\usepackage{cleveref}
\usepackage{booktabs}
\usepackage{algorithm}
\usepackage{todonotes}

\newcommand{\figref}[1]{\figurename~\ref{#1}}
\newcommand{\Hu}{\mathbf{H}}
\newcommand{\R}{\mathbf{R}}
\newcommand{\AH}{\mathcal{A}^H}
\newcommand{\AR}{\mathcal{A}^R}
\newcommand\blfootnote[1]{%
  \begingroup
  \renewcommand\thefootnote{}\footnote{#1}%
  \addtocounter{footnote}{-1}%
  \endgroup
}

\newtheorem{thm}{Theorem}

\usepackage{hyperref}

\usepackage[accepted]{icml2018}

\icmltitlerunning{An Efficient, Generalized Bellman Update For Cooperative Inverse Reinforcement Learning}

\begin{document}

\twocolumn[
\icmltitle{An Efficient, Generalized Bellman Update For\\Cooperative Inverse Reinforcement Learning}

\icmlsetsymbol{equal}{*}

\begin{icmlauthorlist}
\icmlauthor{Dhruv Malik}{equal,berk}
\icmlauthor{Malayandi Palaniappan}{equal,berk}
\icmlauthor{Jaime F.~Fisac}{berk}
\icmlauthor{Dylan Hadfield-Menell}{berk}
\icmlauthor{Stuart Russell}{berk}
\icmlauthor{Anca D.~Dragan}{berk}
\end{icmlauthorlist}

\icmlaffiliation{berk}{Department of Electrical Engineering and Computer Sciences, University of California, Berkeley}

\icmlcorrespondingauthor{Dhruv Malik}{dhruvmalik@berkeley.edu}
\icmlcorrespondingauthor{Malayandi Palaniappan}{malayandi@berkeley.edu}

\icmlkeywords{Value Alignment, AI Safety, Human-Robot Cooperation}

\vskip 0.3in
]

\printAffiliationsAndNotice{\icmlEqualContribution}

\begin{abstract}

Our goal is for AI systems to correctly identify and act according to their human user's objectives. Cooperative Inverse Reinforcement Learning (CIRL) formalizes this \emph{value alignment} problem as a two-player game between a human and robot, in which only the human knows the parameters of the reward function: the robot needs to learn them as the interaction unfolds. Previous work showed that CIRL can be solved as a POMDP, but with an action space size exponential in the size of the reward parameter space. In this work, we exploit a specific property of CIRL---the human is a full information agent---to derive an optimality-preserving modification to the standard Bellman update; this reduces the complexity of the problem by an exponential factor and allows us to relax CIRL's assumption of human rationality. We apply this update to a variety of POMDP solvers and find that it enables us to scale CIRL to non-trivial problems, with larger reward parameter spaces, and larger action spaces for both robot and human. In solutions to these larger problems, the human exhibits pedagogic (teaching) behavior, while the robot interprets it as such and attains higher value for the human.
\end{abstract}

\section{Introduction}\label{Intro}

As AI agents improve in their ability to optimize for a given objective, it becomes increasingly important that these agents pursue the \emph{right} objective. The \emph{value alignment} problem \cite{hadfield2016cooperative,Bostrom:2014:SPD:2678074} is that of ensuring that robots optimize for what people want---that robot objectives are aligned with their end-users' objectives. (We henceforth use robot to refer generically to an AI agent.)

A highly-capable autonomous agent working towards the wrong goal can cause undesired effects, the magnitude of which will tend to increase with the capabilities of the agent. Unfortunately, we humans have a hard time specifying what it is that we actually want. For example, customers may give mistaken instructions to an AI system and system designers may select simple, but potentially incorrect, reward functions to optimize~\cite{faulty}. Optimizing for the wrong objective can lead to unintended and negative consequences~\cite{DBLP:journals/corr/AmodeiOSCSM16}.

\begin{figure}[t!]
\begin{center}
\centerline{\includegraphics[width=\columnwidth]{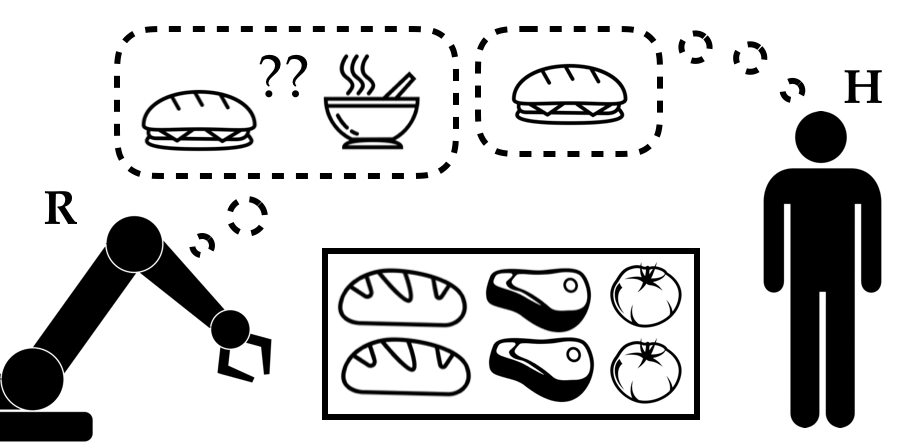}}
\caption{A CIRL game. The human \textbf{H} and the robot \textbf{R} need to work together to prepare a meal. \textbf{R} starts off unaware of which meal \textbf{H} wants, but both \textbf{H} and \textbf{R} get rewarded only if they prepare \textbf{H}'s desired meal. Solving such a CIRL game has thus far been intractable. In Section \ref{bellman}, we derive a modified Bellman update for computing optimal solutions to CIRL games that achieves an exponential reduction in running time and relaxes CIRL's assumption of human rationality.
\vspace{-3.2em}}
\label{value_alignment}
\end{center}
\end{figure} 

Rather than optimize a pre-specified reward function, a robot may instead attempt to infer what people \emph{internally} want but cannot perfectly explicate. The robot can use a person's actions to learn about the reward function over time. The most common approach for this is Inverse Reinforcement Learning (IRL) \cite{ng2000algorithms}. IRL makes two implicit assumptions: 1) that the robot is a passive observer, watching the human, and 2) that the human acts as an expert in isolation, ignoring that the robot needs to learn. 

Cooperative Inverse Reinforcement Learning (CIRL) \cite{hadfield2016cooperative} relaxes these two assumptions. It proposes a formulation in which the human $\Hu$ and the robot $\R$  are on the same team and collaborate to achieve the same goal. CIRL is a two-player game between $\Hu$ and $\R$, in which \emph{both take actions}, and \emph{both get rewarded according to the same reward function}. The key to CIRL is that only $\Hu$ knows the parameters of this reward function. 

Take for instance the domain from \figref{value_alignment}. $\Hu$ and $\R$ work to prepare a meal using three ingredient types: bread, meat, and tomatoes. $\Hu$ wants to prepare either a sandwich (2 bread, 1 meat, 0 tomatoes), or tomato soup (1 bread, 1 meat, 2 tomatoes). $\R$ does not know \emph{a priori} which meal $\Hu$ wants, and, to emulate the difficulty that people have in specifying what they want, we assume $\Hu$ cannot explicate this information directly to $\R$. At every time step, $\R$ and $\Hu$ each prepare a single unit of any ingredient, or no ingredient at all. They both receive reward of 1 if they succeed in preparing the right recipe, and 0 otherwise. 

In this domain, CIRL captures that the human has an incentive for the robot to infer the correct recipe; and that the robot can take actions in response to the human's, as opposed to passively waiting until it knows which recipe is right. Crucially, the robot shares the reward function and has an incentive to maximize the human's internal reward. This creates an incentive to mitigate and avoid unintended consequences from misspecified rewards.

Solving a CIRL game, however, amounts to solving a Dec-POMDP. Previous work has shown that a CIRL game can be reduced to a POMDP. However, the action space in this POMDP is exponential in the size of the reward parameter space. Since POMDP algorithms scale poorly with the size of the action space, non-trivial CIRL games remain difficult to solve with this approach. Additionally, solutions to CIRL are only optimal under the assumption that the human is optimal. This is a strong assumption: it is a well-established fact in cognitive science that humans are often sub-optimal in decision making \cite{tversky1975judgment,simon1957models}. Our contributions in this paper are three-fold:

\noindent\textbf{1. A Modified Bellman Update:} We exploit the fact that the human is a full information agent in CIRL to derive an optimality-preserving modification of the standard Bellman update. This reduces the complexity of the problem by an exponential factor. We show how to apply this modification to existing POMDP solvers (both exact and approximate).

We further show that our modified Bellman update allows us to relax CIRL's assumption of human rationality. We instead only require that the human's policy be parameterized by her Q-values.
This allows us to solve more realistic CIRL games where the human is modelled as sub-optimal. 

\noindent\textbf{2. Empirical Comparison:} We show empirically that our method helps scale POMDP solvers to CIRL games with larger reward parameter and action spaces. We find a speed-up of several orders of magnitude for exact methods, and substantial improvements in value for approximate methods. 

\noindent\textbf{3. Implications:} With the ability to solve more complex CIRL problems, we analyze the solutions that emerge. In contrast to IRL, we see solutions that exhibit implicit communication. The human takes explicitly suboptimal actions that are better signals for the right reward, and the robot attains higher value for the human because it can take advantage of these signals to learn faster. The coordination that emerges is a consequence of the human and robot being on the same team and reasoning about helping each other.

\section{Background} \label{background}

\subsection{POMDPs}

POMDPs provide a rich model for planning under uncertainty \cite{sondik71,Kaelbling:1998:PAP:1643275.1643301}. Formally, a POMDP is a tuple $\langle X, A, Z, T, O, r, \gamma\rangle$ where $X$ is the set of states; $A$ is the set of the agent's actions; $Z$ is the set of observations; $T(x_t, a_t, x_{t+1})$ is the transition distribution; $O(x_{t+1},a_t,z_{t+1})$ is the observation distribution; $r$ is the reward function; and $\gamma$ is the discount factor. 

Consider a simplified instance of the cooking task from \figref{value_alignment} where $\Hu$ picks her actions according to only her desired recipe and the quantity of each ingredient prepared so far, i.e., she does not consider $\R$'s past or future behavior when picking her actions. The simplified cooking task is now a POMDP: $\R$ is the agent and $\Hu$ is a part of the environment. The state specifies $\Hu$'s desired recipe and the quantity of each ingredient already prepared. Thus, $\Hu$ picks her actions as a function of only the state.

In a POMDP, the agent cannot observe the state; instead, it maintains a belief $b\in \Delta X$, where $b(x)$ is the probability that the agent is in state $x$. At each time step, the agent receives an observation that helps inform its decisions. The agent in our cooking task, $\R$, does not know $\Hu$'s desired recipe---a component of the state. $\R$ observes $\Hu$'s actions and tries to infer the desired recipe from $\Hu$'s behavior.

The behavior of the agent is specified by a conditional plan $\sigma = (a, v)$; $a$ denotes the agent's action and $v$ is a mapping from observations to future conditional plans for the agent to follow. An example conditional plan for $\R$ is: prepare meat now and if $\Hu$ responds by preparing bread, prepare a second slice of bread; if $\Hu$ prepares tomatoes, prepare another batch of tomatoes; or if $\Hu$ prepares meat, do not prepare any ingredient.

The $\alpha$-vector of a conditional plan contains the value of following the plan at any given state: \begin{align}
    \alpha_{\sigma}(x)  &=R(x) + \gamma\sum_{x'\in X}\sum_{z\in Z}P(x', z\mid x, a)\alpha_{v(z)}(x')\label{alpha_basic}
\end{align}
The value of a plan at a belief $b$ is the expected value of the plan across the states i.e. $V_{\sigma}(b) = b \cdot \alpha_{\sigma} = \sum_{x\in X} b(x)\alpha_{\sigma}(x)$. The goal of an agent in a POMDP is to find the plan with maximal value from her current belief. 

Value iteration \cite{sondik71} can be used to compute the optimal conditional plan. This algorithm starts at the horizon and works backwards. It generates new conditional plans at each time-step and evaluates them according to Eq. 1. It constructs all potentially optimal plans of length $T$ and selects the one with maximal value at the initial belief.

\subsection{Cooperative Inverse Reinforcement Learning}

Now, consider an instance of the cooking task where $\Hu$ is a second agent in the game and no longer behaves independently of $\R$. There is now a strong interdependence between $\Hu$'s and $\R$'s behavior: $\Hu$'s actions both depend on and influence $\R$'s belief. This problem is now no longer a POMDP; it is a CIRL game.

A CIRL game is an asymmetric-information two-player game between a human $\Hu$ and a robot $\R$ ~\cite{hadfield2016cooperative}. $\Hu$ knows the true reward function and $\R$ does not initially.  Formally, a CIRL game is a tuple: $M = \langle X, \{\AH, \AR\}, T, \{\Theta, r\}, \gamma\rangle$. $X$ is the set of observable world-states; $\AH$ and $\AR$ are the actions available to $\Hu$ and $\R$ respectively; $T(x_t, a^H_t, a^R_t, x_{t+1})$ is the transition distribution; $\Theta$ is the set of reward parameters; $r$ is the parameterized reward function shared by both agents; $\gamma$ is the discount factor. A solution to a CIRL game is a pair of policies---one for $\Hu$ and $\R$ each---that maximizes the expected reward obtained by $\Hu$ and $\R$.

In our cooking task, $\Theta$ is the set of possible recipes. $\R$ does not know $\Hu$'s desired recipe, $\theta\in \Theta$. The reward function $r$ is parameterized by $\Theta$: both agents receive a reward of 1 if they succeed in preparing $\Hu$'s desired recipe.

\noindent\textbf{Reducing a CIRL game to a POMDP} A CIRL game is a Dec-POMDP \cite{BGIZmor02} but it can be reduced to a POMDP where the optimal policy corresponds to optimal CIRL policy pairs \cite{hadfield2016cooperative}. The states in this POMDP are tuples of world-state and reward parameter: $S = X\times \Theta$; the actions are tuples ($\delta^H$, $a^R$) specifying a decision rule $\delta^H: \Theta \rightarrow \AH$ for $\Hu$, which maps reward parameters to human actions, and an action $a^R$ for $\R$; the observations are $\Hu$'s action at the last time step.

An example action in the reduced POMDP of the cooking task is a tuple, where the first entry specifies that $\Hu$ prepares bread if she prefers a sandwich and prepares tomatoes if she prefers soup, and the second entry specifies that $\R$ prepares bread (regardless of its belief).

\begin{figure}[]
\vskip 0.2in
\begin{center}
\centerline{\includegraphics[width=\columnwidth]{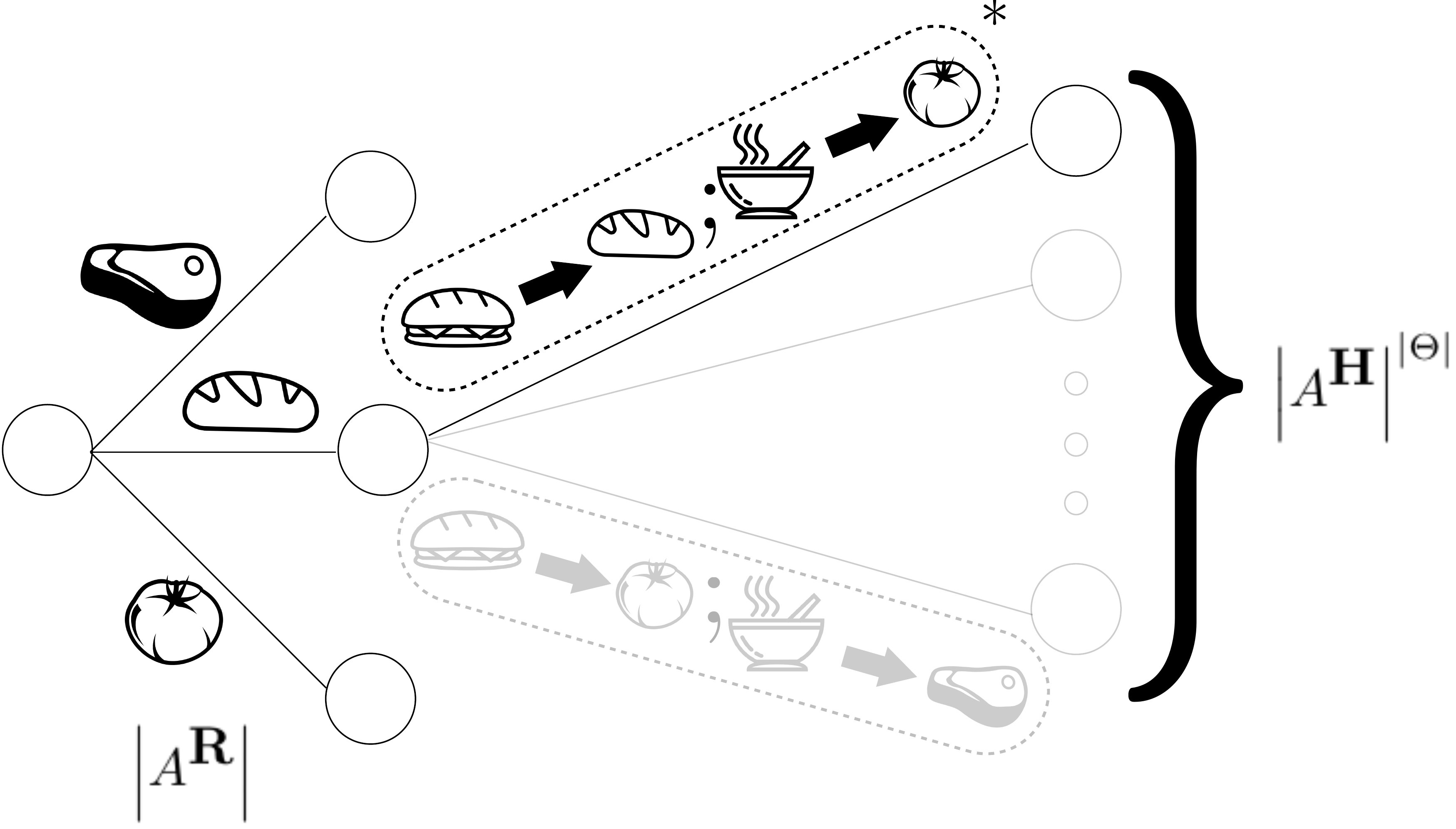}}
\caption{A node in the search tree from the POMDP reduction of our example CIRL game. Actions are tuples that contain an action for $\R$ and a decision rule for $\Hu$ -- a mapping from her desired recipe to an action. This leads to a branching factor of $|\AH|^{|\Theta|}|\AR|$ and makes application of POMDP methods inefficient. In Section \ref{derivation}, we derive a modified Bellman update that prunes away all of H's decision rules but the optimal response. (In the diagram, the gray branches are pruned away by our modified Bellman update.)}
\label{cirl_tree}
\end{center}
\vskip -0.4in
\end{figure}

This reduction enables us to solve a CIRL game using POMDP algorithms. However, the size of the action space in this POMDP is $|\AH|^{|\Theta|}|\AR|$, as shown in \figref{cirl_tree}. (There are $|\AH|^{|\Theta|}$ possible decision rules for $\Hu$ and $|\AR|$ actions for $\R$.) In other words, the action space in this POMDP grows exponentially with the size of the reward parameter space. Exact POMDP algorithms are exponential in the size of the action space, so this approach can only be applied to very small CIRL problems.

Additionally, the policy for $\R$ that is output by the reduced POMDP is optimal only if $\Hu$ is perfectly rational, i.e., if $\Hu$ is guaranteed to always pick the optimal action. This is an unrealistic assumption: humans are not idealized rational agents \cite{tversky1975judgment, simon1957models}. 

\section{A Modified Bellman Update for CIRL} \label{bellman}

If $\Hu$ were following a fixed policy based on the state $s~=~(x,~\theta)$, we could encode $\Hu$ as a part of the environment. However, in the interactive setting of a CIRL game, $\Hu$ may plan for changes in $\R$'s belief. If we encode $\Hu$ in the environment, the dynamics change in response to $\R$'s intended plan and the problem is no longer a POMDP. Our main contribution is to derive a modified Bellman update for POMDP algorithms to solve this problem. 

Our key idea is as follows. During planning, we know $\R$'s intended future response to each of $\Hu$'s actions. $\Hu$ has full state information, so the $\alpha$-vectors in value iteration allow us to directly compute $\Hu$'s Q-values. We can therefore also compute her optimal action based on $\R$'s intended future response. This means we do not have to reason over the set of decision rules for $\Hu$: we can solve a CIRL game by instead solving a POMDP with time-varying dynamics and, importantly, where the action space has size $|\AR|$. This is exponentially smaller than the action space of size $|\AH|^{|\Theta|}|\AR|$ in the reduced POMDP. This amounts to a modified Bellman update.

\subsection{The Transition Dynamics}

If $\Hu$ follows a policy that depends only on the state $s = (x, \theta)$, the dynamics of the game can be computed as:
\vspace{-0.05in}
\begin{align*}
    &P(s', a^H\mid s, a^R) = P((x', \theta'), a^H\mid (x, \theta), a^R)\\ 
    &= P((x', \theta')\mid (x, \theta), a^H, a^R)\cdot P(a^H\mid (x, \theta), a^R)\\
    &= T(x, a^H, a^R, x')\cdot \mathds{1}(\theta = \theta') \cdot P(a^H\mid x, a^R, \theta)\\
    &\stackrel{a.}= T(s, a^H, a^R, s')\cdot P(a^H\mid x, a^R, \theta)
\end{align*}
\blfootnote{a. We let $T(s, a^H, a^R, s') = T(x, a^H, a^R, x')\cdot \mathds{1}(\theta = \theta')$.}However, in the CIRL formulation, $\Hu$ does not behave according to a fixed policy. $\Hu$, who is assumed to be rational, behaves according to her Q-values and picks the action that maximizes her expected value. Due to the interdependence between $\Hu$'s and $\R$'s behavior, these Q-values depend on $\R$'s conditional plan. The dynamics then are:

\begin{align}
   &P(s', a^H\mid s, \sigma) = P((x', \theta'), a^H\mid (x, \theta), (a^R, v))\nonumber\\
    &= T(x, a^H, a^R, x')  \cdot \mathds{1}(\theta' = \theta)\cdot P(a^H\mid x, a^R, v, \theta)\nonumber\\
    &= T(s, a^H, a^R, s') \cdot \mathds{1}(a^H = \arg\max_{a^H}Q_{H}(s, a^H, 
    \sigma))\label{alpha_vi}
\end{align}
These dynamics change over time since they depend on the robot's future behavior. However, $\R$'s behavior depends on these dynamics, so, we cannot pre-compute them as part of a POMDP reduction. However, we do have access to $\R$'s future conditional plan \emph{during} planning. This means we can compute $\Hu$'s Q-values, and, consequently, the transition probabilities, with a modification to the Bellman update.

\subsection{Adapting POMDP Value Iteration}\label{derivation}

POMDP value iteration rolls back the values of the game from the horizon, storing them as $\alpha$-vectors.  If $\R$ follows a plan $\sigma = (a^R, v)$, then we can compute $\Hu$'s Q-values as $$Q_{H}(s, a^H, \sigma) = \sum_{s'} T(s, a^H, a^R, s') \cdot \alpha_{v(a^H)}(s').$$ 
$\Hu$'s optimal action maximizes this expression. To leverage this, we adapt the Bellman update in Eq. 1 to replace the dynamics of the game with $P(s', a^H\mid s, \sigma)$ from Eq. 2. The modified Bellman update is then: \begin{align}
    \alpha_{\sigma}(s)  &=R(s) + \gamma\cdot \max_{a^H}\sum_{s'\in S}T(s,a^H, a^R, s')\cdot \alpha_{v(a^H)}(s').\label{bellman_update}
\end{align}
We can then use value iteration with this modified Bellman update to compute the $\R$'s optimal policy. The following theorem establishes that the modified Bellman update, Eq. 3, is optimality-preserving.
\begin{thm}\label{optimal}For any CIRL game, the policy computed by value iteration with the modified Bellman update is optimal.
\end{thm}
All theorem proofs are presented in Appendix A.

This modification to the Bellman update allows us to solve a CIRL game without having to include the set of $\Hu$'s decision rules in the action space. As depicted in \figref{cirl_tree}, the modified Bellman update computes $\Hu$'s optimal action given the current state and the robot's plan; all of $\Hu$'s other actions are pruned away in the search tree. The size of the action space is then $|\AR|$ instead of $|\AH|^{|\Theta|}|\AR|$. POMDP algorithms are exponential in the size of the action space; this modification therefore allows us to solve CIRL games much more efficiently. The following theorem establishes the complexity gains made by algorithm.
\begin{thm}\label{exact}The modification to the Bellman update presented above reduces the time and space complexity of a single step of value iteration by a factor of $\mathcal{O}\left(|\AH|^{|\Theta|}\right)$.
\end{thm}

\subsection{Relaxing CIRL's Assumption of Rationality}

To achieve value alignment, we can now efficiently solve a CIRL game to find an optimal policy for $\R$. However, this policy is optimal only if $\Hu$ is perfectly rational: a strong assumption. This is rarely true in reality; we thus want to find an optimal policy for $\R$ even when $\Hu$ is sub-optimal.

In addition to improving efficiency, our modified Bellman update allows us to do exactly that and relax CIRL's assumption of rationality. The dynamics of our modified Bellman update, presented above as Eq. 2, do not require that $\Hu$ is perfectly rational. These dynamics will remain well-defined so long as we know the distribution over $\Hu$'s actions, $\pi_H$, and can compute the probability that she picks any action from her current state. To avoid compromising the interactive nature of CIRL, we require that $\pi_H$ must be a function of $\Hu$'s Q-values, which account for the robot's future behavior. The dynamics of the game are then given by:$$P(s', a^H\mid s, \sigma) = T(s, a^H, a^R, s')\cdot \pi_H(a^H\mid Q_H(s, a^H, \sigma)).$$ 
The modified Bellman update is then:\begin{align}
    &\alpha_{\sigma}(s)  =R(s) +\gamma\cdot \sum_{a^H}\pi_H(a^H\mid Q_H(s, a^H, \sigma))\cdot\nonumber\\
    &\ \ \ \ \sum_{s'\in S}T(s,a^H, a^R, s')\cdot  \alpha_{v(a^H)}(s').\label{bellman_advanced_update}
\end{align}
With this modified Bellman update, we may now use value iteration to solve CIRL games without assuming that the human is perfectly rational. We instead only require that the human selects her actions with respect to her Q-values. This restriction is rather broad and includes a variety of models of human decision making from cognitive science. A popular example of such a model is Boltzmann-rationality, where the human picks her actions according to a Boltzmann distribution over her Q-values, i.e., $$\pi_H(a^H\mid Q_H(s, a^H, \sigma)) \propto \exp(\beta\cdot Q_H(s, a^H, \sigma))$$where $\beta$ is a parameter which controls how rational the human is. (A higher $\beta$ corresponds to a more rational human.)
\begin{algorithm}[t!]
\caption{Adapted Value Iteration for CIRL Games}\label{exactvi}
\begin{algorithmic}[1]

\STATE $\Gamma_{t} \gets$ Set of trivial plans
    \FOR{$t \in \{T-1, T-2, \ldots, 1, 0\}$}
        \STATE $\Gamma_{t+1} \gets \Gamma_{t}$
        \STATE{$\Gamma_t \gets$ Set of all plans beginning at time t}
        \FOR{$\sigma \in \Gamma_t$}
            \FOR{$s = (x, \theta) \in S$}
                \STATE $Q_{H}(s, a^H, \sigma) = \sum_{s'} T(s, a^H, a^R, s') \cdot$
                \STATE $\ \ \ \ \alpha_{v(a^H)}(s')$
                \STATE $\alpha_{\sigma}(s) = R(s) + \gamma\cdot \sum_{a^H}$
                \STATE $\ \ \ \ \pi_H(a^H\mid~Q_H(s, a^H, \sigma)) \cdot Q_{H}(s, a^H, \sigma)$
            \ENDFOR
        \ENDFOR
        \STATE $\Gamma_{t} \gets$ Prune($\Gamma_{t}$)
    \ENDFOR
    \STATE $a^R_* = argmax_{\sigma\in\Gamma_0} \alpha_{\sigma}\cdot b_0$
    \STATE \textbf{Return} $a^R_*$
\end{algorithmic}
\end{algorithm}
\vspace{-1.3em}

The time and space complexity of value iteration with this Bellman update is identical to that with the modified Bellman update presented in Section 3.2, and analyzed in Theorem 2. The pseudocode for our adapted algorithm is presented as Algorithm \ref{exactvi} below.
\section{Adapting Approximate Algorithms} \label{approx_alg}

Approximate algorithms for POMDPs often rely on variants of the Bellman update. This lets us use our modified Bellman update to improve approximate algorithms for CIRL.

\subsection{PBVI}
\noindent\textbf{Background}
Point Based Value Iteration (PBVI) is an approximate algorithm used to solve POMDPs \cite{pineau2003point}. The algorithm maintains a representative set of points in belief space and an $\alpha$-vector at each of these belief points. It performs approximate value backups at each of these belief points using this set of $\alpha$-vectors. Let $\Gamma_{t+1}$ denote the set of $\alpha$-vectors for plans that begin at time $t+1$. The value at time $t$ at a belief $b$ is approximated as:\begin{align*}V(b) &= \max_{a\in A}\Bigg[\sum_{s\in S}R(s)b(s)\\&+ \gamma \sum_{o\in O}\max_{\alpha\in \Gamma_{t+1}}\sum_{s\in S}\left(\sum_{s'\in S}P(s',o\mid s,a)\alpha(s)\right)b(s)\Bigg].\end{align*} The algorithm trades off computation time and solution quality by expanding the set of belief points over time: it randomly simulates forward trajectories in the POMDP to produce new, reachable beliefs.

\noindent\textbf{Our Adaptation}
If $\R$ takes action $a^R$ and follows a conditional plan $\sigma$, then $\Hu$'s Q-values are $Q_{\Hu}(x, a^H, a^R, \alpha) = \sum_{s'} T(s, a^H, a^R, s')\cdot \alpha_{\sigma}(s')$. Notice that we can compute these Q-values at each step of PBVI. This lets us use the modified Bellman update and to adapt PBVI to solve CIRL games specifically. We replace the transition-observation distribution in the PBVI backup rule with
\begin{align*}
   P(s', a^H\mid s, a^R, \alpha) &= T(s, a^H, a^R, s') \cdot \\&\pi_{\Hu}(Q_{\Hu}(x, a^H, a^R, \alpha)).
\end{align*}
The modified backup rule for PBVI is thus given by\begin{align*}V(b) &= \max_{a^R\in \AR}\Bigg[\sum_{s\in S}R(s)b(s)\\+ \gamma &\sum_{a^H} \max_{\alpha\in \Gamma_{t+1}}\sum_{s\in S}\left(\sum_{s'\in S}P(s',a^H\mid s,a^R, \alpha)\alpha(s)\right)b(s)\Bigg]. \end{align*}

We now show that the approximate value function in PBVI converges to the true value function. Let $\epsilon_B$ denote the density of the set of belief points $B$ in PBVI. Formally, $\epsilon_B = \max_{b\in \Delta}\min_{b'\in B} ||b-b'||_1$ is the maximum distance from any reachable, legal belief to the set $B$.

\begin{thm}\label{pbvi_proof}
For any belief set $B$ and horizon $n$, the error of our adapted PBVI algorithm $\eta = ||V_n - V_n^*||_{\infty}$ is bounded as
$$\eta \leq \frac{(R_{\max} - R_{\min})\epsilon_B}{(1-\gamma)^2}.$$
\end{thm}

\subsection{POMCP}
\noindent\textbf{Background}
POMCP is a Monte Carlo tree-search (MCTS) based approximate algorithm for solving large POMDPs \cite{silver2010monte}. The algorithm constructs a search tree of action-observation histories and uses Monte Carlo simulations to estimate the value of each node in the tree. During search, actions within the tree are selected by UCB1. This maintains a balance between exploiting actions known to have good return and exploring actions not yet taken \cite{Kocsis:2006:BBM:2091602.2091633}. At leaf nodes, a rollout policy accrues reward which is then backed up through the tree. The algorithm estimates the belief at each node by keeping track of the hidden state from previous rollouts. 

POMCP scales well with the size of the state space, but not with the size of the action space, which determines the branching factor in the search tree. POMCP is thus ill-suited to solving the reduced POMDP of CIRL games since the size of the action space is $|\AH|^{|\Theta|}|\AR|$.

\noindent\textbf{Our Adaptation}
Using the idea behind our modified Bellman update, we adapt POMCP to solve CIRL games more efficiently. We approximate $\Hu$'s policy while running the algorithm (much like we exactly compute $\Hu$'s policy in exact value iteration). We maintain a live estimate of the sampled Q-values for $\Hu$ at each node. With enough exploration of the search tree (for instance, if actions are selected using UCB1), the estimated Q-values converge to the true values (in the limit). This guarantees that $\Hu$'s policy converges to her true policy. The following result establishes convergence of our algorithm.

\begin{thm}\label{pomcp_proof}
With suitable exploration, the value function constructed by our adapted POMCP algorithm converges in probability to the optimal value function, ${V(h) \rightarrow V^*(h)}$. As the number of visits $N(h)$ approaches infinity, the bias of the value function $\mathbb{E}[V(h) - V^*(h)]$ is $O(log(N(h))/N(h))$.
\end{thm}

The pseudocode for our adapted PBVI and our adapted POMCP algorithm is presented as Algorithm 1 and 2 respectively in Appendix B.

\section{Related Work}

\noindent\textbf{POMDP Algorithms}
We chose to explicate our modified Bellman update in the context of PBVI and POMCP because they are the seminal point-based and MCTS algorithms respectively, for solving POMDPs. For example, SARSOP \cite{Kurniawati08sarsop:efficient} and DESPOT \cite{YeSom17}, two state-of-the-art algorithm for POMDPs, are variants of PBVI and POMCP respectively. The principles we outlined in Sections \ref{bellman} and \ref{approx_alg} can be easily adapted to a large variety of point-based and MCTS algorithms, including any which may be developed in the future, to derive even more efficient algorithms for solving CIRL games.

\noindent\textbf{MOMDP Algorithms} The POMDP representation of CIRL is also a mixed-observability Markov decision process (MOMDP) since the state space can be factored into a fully- and a partially-observable component. This structure allows for more efficient solution methods; \citet{ong2010planning} leverage the factored nature of the state space to create a lower dimensional representation of belief space. This core idea is orthogonal to ours, which exploits CIRL's information asymmetry instead. The two can be leveraged together.

\noindent\textbf{Dec-POMDP Algorithms}
Dec-POMDP algorithms can be used to solve CIRL directly, without using the POMDP reduction. These solution methods are generally intractable, but recent work has made progress on this front. Such Dec-POMDP algorithms which attempt to prune away unreasonable strategies resemble our approach. \citet{conf/aips/AmatoDZ09} use reachability analysis to identify reachable states, then consider all policies which are useful at such states. \citet{Hansen04dynamicprogramming} model other agents’ possible strategies as part of a player’s belief, and prune away weakly dominated strategies at each step. While such approaches use heuristics to prune away some suboptimal strategies, we leverage the information structure of CIRL to compute the optimal strategy for $\Hu$ and prune away \textit{all} other strategies. This guarantees an exponential reduction in complexity while preserving optimality; this is not true for the other methods.

\noindent\textbf{Value Alignment} Recent work has explored relaxing the rationality requirement of CIRL \cite{fisac2017pragmatic}. Our work improves on their relaxation in several ways: (1) Their Bellman update assumes that the human acts Boltzmann-rationally. Our modification can model a large variety of human behaviors (including this). (2) Their discretized belief value iteration algorithm has neither the guarantee of optimality of our adapted VI algorithm nor the scalability of our adapted POMCP/PBVI algorithms. 

\section{Experiments}\label{experiments}

We now verify that our modified Bellman update allows POMDP algorithms to solve CIRL games more efficiently than the standard update. We ran three experiments: one for exact value iteration (VI), PBVI, and POMCP each. The results of the PBVI experiment are presented in Appendix C.1 due to space constraints. To verify the results of our experiments, we ran further experiments on a second domain. The details and results of these experiments are presented in Appendix C.2.

\subsection{Experimental Design}

\noindent\textbf{Domain}
Our experimental domain is based on our running example from Section \ref{Intro}. Assume there are $m$ recipes and $n$ ingredients. The state space is an $n$-tuple representing the quantity of each ingredient prepared thus far. At each time step, each agent can prepare any of the $n$ ingredients or none at all. Each of the $m$ recipes corresponds to a different $\theta$ (i.e. reward parameter) value. Both agents receive a reward of $1$ if $\Hu$'s desired recipe is prepared correctly and a reward of $0$ otherwise. The robot $\R$ begins the game entirely uncertain about $\Hu$'s desired recipe i.e. $\R$ has a uniform belief over $\Theta$.

We want to stress that this experimental domain is not trivial: one of the domains we managed to solve in our experiments had $\sim 10^{10}$ states.
\begin{table}[]
\caption{Time taken (s) to find the optimal robot policy using exact VI and our adaptation of it for various numbers of possible recipes. NA denotes that the algorithm failed to solve the problem.}
\label{vi-table}
\begin{center}
\begin{small}
\begin{sc}
\begin{tabular}{lcccr}
\hline
\abovespace\belowspace
$\#$ Recipes & Exact VI & Ours \\
\hline
\abovespace
2    & 4.448 $\pm$ 0.057 & 0.071 $\pm$ 0.013\\
3   & 394.546 $\pm$ 6.396 & 0.111 $\pm$ 0.013\\
4    & NA& 0.158 $\pm$ 0.003 \\
5    & NA&0.219 $\pm$ 0.007 \\
\belowspace
6     & NA&0.307 $\pm$ 0.005\\
\hline
\end{tabular}
\end{sc}
\end{small}
\end{center}
\vskip -0.3in
\end{table}

\begin{figure*}[t!]
\vskip 0.2in
\begin{center}
\centerline{\includegraphics[width=\textwidth]{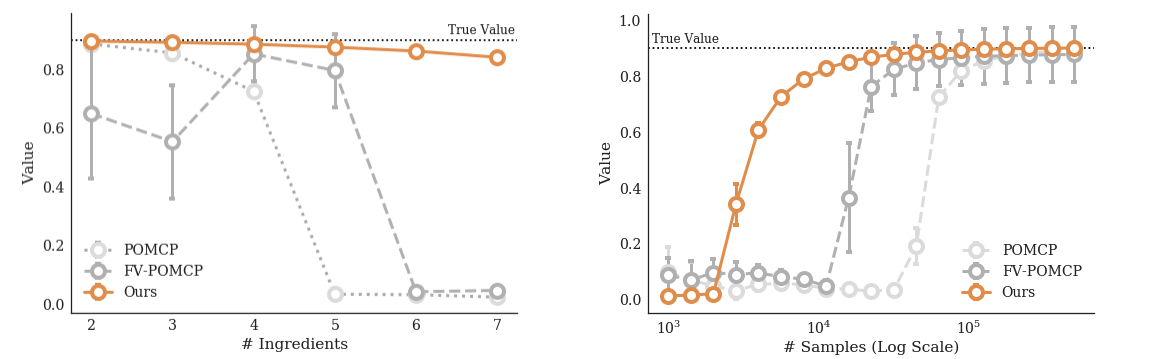}}
\caption{(Left) The value attained by POMCP, FV-POMCP, and our adapted algorithm in 30,000 samples with various numbers of ingredients. (Right) Value attained by POMCP, FV-POMCP and our approximate algorithm with 2 recipes and 6 ingredients.\vspace{-1em}}
\label{pomcp}
\end{center}
\vskip -0.2in
\end{figure*}

\noindent\textbf{Manipulated Variables} Our primary variable is the type of Bellman update used: modified vs. standard. We also varied the number of recipes, i.e., size of the reward parameter space, and the number of ingredients, i.e., size of $\Hu$'s and $\R$'s action space.

\noindent\textbf{Dependent Measures} In our first experiment (exact VI), we measured the time taken by the algorithms to solve the problem.
In our second experiment (PBVI), we measured the value attained by the algorithms in a fixed time. In our third experiment (POMCP), we measured the value attained by the algorithms in a fixed number of samples.

\noindent\textbf{Hypothesis} \textit{POMDP algorithms are more efficient at solving CIRL games with the modified Bellman update than with the standard Bellman update.}

\subsection{Analysis}

\noindent\textbf{Exact VI} In our first experiment, we compared the time taken by exact VI and by our adaptation of it with the modified Bellman update. We first fixed the number of ingredients at two and varied the number of recipes in the domain.  Table \ref{vi-table} compares the results. For the simpler problems, where the number of recipes was 2 or 3, our adapted algorithm solved the problem up to $\sim$3500$\times$ faster than exact VI. On more complex problems where the number of recipes is greater than 3, exact VI failed to solve the problem after depleting our system's 16GB memory; in contrast, our adapted algorithm solved each of these more complex problems in less than 0.5 seconds. We next fixed the number of recipes and compared the performance of both these algorithms for various numbers of ingredients. Both the exact methods, but especially the one using the standard Bellman update, scaled much worse with the number of ingredients than with the number of recipes. With even three ingredients, exact VI timed out and failed to solve the problem within two hours; our algorithm however solved the problem in five seconds.

\noindent\textbf{POMCP} We compared the value attained in 30,000 samples by POMCP and by our adaptation with the modified Bellman update. We additionally compared these algorithms with FV-POMCP, a state-of-the-art MCTS method for solving MPOMDPs, a type of Dec-POMDP in which all agents can observe each others' behavior (as in CIRL).

We first fixed the number of recipes at 2 and varied the number of ingredients. Our adapted algorithm outperformed the other two algorithms across the board, especially for large numbers of ingredients. The results of this comparison are presented in \figref{pomcp} (left). POMCP did poorly on games with more than 4 ingredients. Although FV-POMCP scaled better to more complex games than POMCP, its values had high variance. For the largest games, with 6 and 7 ingredients, our adapted algorithm was the only one capable of solving the problem in 30,000 iterations. We also compared the value attained by each algorithm across 500,000 samples on the 6 ingredient game. The results of this comparison are depicted in \figref{pomcp} (right). Our algorithm converged to the true value faster than either of the other algorithms.

We next fixed the number of ingredients at 4 and varied the number of recipes. We found that the results of this experiment broadly matched the results of our previous experiment where we varied the number of ingredients. For example, with 4 recipes, our method achieves an average value of $0.631 \pm 0.221$ in 30,000 iterations while POMCP gets $0.429 \pm 0.183$ and FV-POMCP gets $0.511 \pm 0.124$. 

These results together demonstrate that POMDP algorithms with our modified Bellman update scales much better to more complex CIRL games than with the standard Bellman update. This offers strong evidence for our hypothesis.

\section{Discussion}

The previous section showed that we can now solve larger, non-trivial CIRL games. While we are still far from addressing value alignment in the high dimensional and continuous real world, our work allows us to analyze CIRL solutions to non-trivial problems and understand their implications for value alignment.

\subsection{CIRL vs IRL}

In the absence of CIRL solutions, a standard approach to learning the human's internal reward is Inverse Reinforcement Learning (IRL) \cite{ng2000algorithms}. We thus compare what advantages CIRL has compared to IRL.  On a collaborative task, IRL is equivalent to assuming that $\Hu$ chooses her actions in isolation, and $\R$ uses observations of her behavior to infer her preferences. Specifically, $\Hu$ solves a single-agent, fully-observable, variant of the problem, and $\R$ responds by solving the POMDP described in Section~\ref{background}.

\begin{figure}[t!]
\begin{center}
\centerline{\includegraphics[width=\columnwidth]{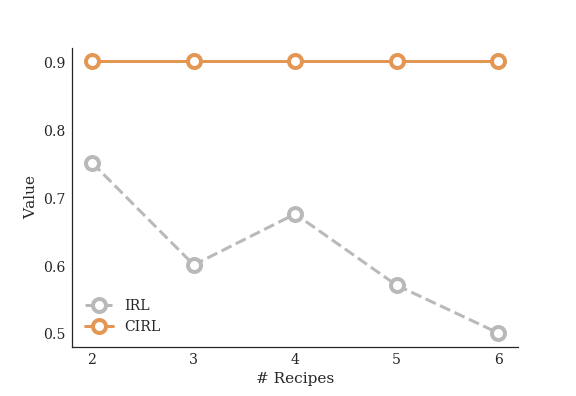}}
\caption{Value attained by CIRL and standard IRL on the cooking domain with various numbers of possible recipes. Unlike IRL, CIRL produces solutions where $\Hu$ picks her actions pedagogically and $\R$ reasons about $\Hu$ accordingly.\vspace{-2em}}
\label{irlcomp}
\end{center}
\vskip -0.2in
\end{figure}

We fix the number of ingredients at 3 and vary the number of recipes. \figref{irlcomp} shows the results. In each experiment the optimal CIRL solution prepares the correct recipe while the IRL solution fails to do so up to 50\% of the time. To understand the nature of this difference in performance, we analyze the CIRL and IRL solutions. Consider a case of our running example with two recipes. The state is a tuple $(\#meat, \#bread, \#tomatoes)$ and $\Theta = \{sandwich = (1,2,0), soup =  (1,1,2)\}$. For both approaches, $\R$ initially prepares meat. In the baseline IRL solution, $\Hu$ can initially make any ingredient if she wants soup and can make meat or bread if she wants a sandwich. In each case, she chooses uniformly at random between allowed ingredients. This conveys some information about her desired recipe, but is not enough to uniquely identify it. Since the same ingredient is optimal for multiple recipes, $\R$ is still confused after one turn. This means $\R$ will sometimes fail to complete the desired recipe, reducing average utility.

The CIRL solution, in contrast, relies on the implicit communication between the human and the robot. Here, if $\Hu$ wants soup, she prepares tomatoes, as opposed to any ingredients that are common with the sandwich. Even more interestingly, she waits (i.e. does nothing) if she wants a sandwich. This is pure signaling behavior---\emph{waiting is suboptimal in isolation}, but picking an ingredient is more likely to confuse the robot. In turn $\R$ knows that $\Hu$ would have picked tomatoes if she wanted soup, and responds appropriately.

In other words, $\Hu$ teaches the robot about her preferences with her action selection. This works because $\Hu$ knows that $\R$ will interpret her behavior pragmatically, i.e., $\R$ expects to be taught by $\Hu$. This is reflected in the experiment: the optimal CIRL solution prepares the correct recipe each time.

The value alignment problem is necessarily cooperative: without the robot, the human is unable to complete her desired task, and without explicit signaling from the human, the robot learns inefficiently, is less valuable and is more likely to make a mistake. Pedagogic behavior from $\Hu$ naturally falls out of the CIRL solution. In response, $\R$ interprets $\Hu$'s actions pragmatically. These instructive and communicative behaviors allow for faster learning and create an opportunity to generate higher value for the human.

\subsection{CIRL with Suboptimal Humans}

To further investigate the performance of CIRL in realistic settings (e.g., where $\Hu$ may not be rational), we ran another experiment. We varied whether $\Hu$ behaved according to CIRL or IRL, $\R$'s model of $\Hu$ in training (rational or Boltzmann-rational), and the actual model of $\Hu$ (same as previous). We measured the proportion of times they prepared the correct recipe in each setting, fixing the number of ingredients at 3 and recipes at 4. \figref{rational_comp} in Appendix D shows the results. (We also conducted a more comprehensive experiment with 20 human behaviors instead of 2. Results are presented in Appendix E.)

Averaged across different models of $\Hu$ used to train $\R$, when $\Hu$ behaved according to CIRL, $\Hu$ and $\R$ succeeded in preparing the correct recipe $>90\%$ of the time. This was also true when $\Hu$ behaved Boltzmann-rationally. This suggests that the pedagogic behavior that arises from CIRL makes it more robust to any sub-optimality from $\Hu$. In contrast, when $\Hu$ behaved as in IRL  (i.e., not pedagogically), they only prepared the correct recipe $\sim$70$\%$ of the time when $\Hu$ was rational, and $\sim$40$\%$ of the time when $\Hu$ was not. So, the importance of pedagogic behavior from $\Hu$ to achieve value alignment is clear.

\subsection{Do People Behave Pedagogically?}

A question arises at this point as to whether \emph{real} people will adopt the pedagogic behavior predicted by CIRL solutions. To start testing this, we ran a (very preliminary) pilot study to start investigating whether CIRL improves interactions with real people. The details of this experiment are presented in Appendix F. We found some encouraging evidence that suggests people do indeed behave pedagogically when collaborating with a robot; and that a CIRL-trained robot is can be better at exploiting this pedagogic behavior to achieve fluid human-robot collaboration than an IRL-trained robot.

In future work, we plan to conduct a more extensive human subjects study to validate these preliminary findings. We additionally plan to explore techniques to make the robot better elicit pedagogic behavior from the human in their interaction and to make CIRL robust to variations in human behavior.

\section*{Appendix}
\appendix
\section{Proofs}
In this section, we present the proofs for the propositions and theorems in the main paper.

\textbf{Theorem 1.} For any CIRL game, the policy computed by value iteration with the modified Bellman update is optimal.
\begin{proof}
During POMDP value iteration, values are propagated from the horizon in the form of $\alpha$ vectors, which store the expected values of executing a conditonal plan from each state. In CIRL games, $\Hu$ is a full information actor, which means she has knowledge of her true reward parameter $\theta$. So, given $\R$'s conditional plan, she can directly compute her Q values from the $\alpha$ vectors. The presence of the inner $\max$ in our modified update rule ensures that $\Hu$ always picks the action corresponding to her maximum Q value. This implies that at every backup step, our modified Bellman update never prunes away the optimal action. Hence, the output policy, which is the best policy among those not pruned away, must be optimal.
\end{proof}
\vskip 0.2in
\textbf{Theorem 2.} The modification to the Bellman update presented above reduces the time and space complexity of a single step of value iteration by a factor of $\mathcal{O}\left(|\AH|^{|\Theta|}\right)$.
\begin{proof}
The complexity of one step of POMDP value iteration is linear in the size of the action space, $|A|$ \cite{russell1995modern}. Since the structure of our algorithm is identical to that of exact value iteration, this is also true for our algorithm.
    
The action space in the POMDP reduction of CIRL has size $|\AH|^{|\Theta|}|\AR|$. Our modification to the Bellman update reduces the size of the action space to simply $|\AR|$. Therefore, our algorithm reduces the time taken to run, and number of plans generated at, each time step by a factor of $|\AH|^{|\Theta|}$.
\end{proof}
\vskip 0.2in
\textbf{Theorem 3.} For any belief set $B$ and horizon $n$, the error of our adapted PBVI algorithm $\eta = ||V_n - V_n^*||_{\infty}$ is bounded as
$$\eta \leq \frac{(R_{\max} - R_{\min})\epsilon_B}{(1-\gamma)^2}.$$
\begin{proof}
    Since the dynamics of our problem are now time-varying instead of static, the backup operator applied at every time-step changes. Let $H_t$ denote the backup operator applied to compute the value of $V_t$. It will suffice to show that each such backup operator $H_t$ is a contraction mapping. The result then follows by following the proof of Theorem 1 in \cite{pineau2003point} exactly. We will prove the result by showing that each $H_t$ is the backup operator for a specific POMDP and thus, for this POMDP's corresponding belief MDP; it must therefore be a contraction mapping.
    
    Take $H_t$ for some $1 \leq t\leq n$. We will now construct a new POMDP for which $H_t$ is the backup operator. Let $\hat{S} = S \times \Gamma_{t+1}$, where $\Gamma_{t+1}$ denotes the set of $\alpha$-vectors from our original problem at time $t+1$. Let the $\alpha$-vector component of the state be static i.e. $P((s', \alpha')\mid (s,\alpha), a^R) = 0$ if $\alpha \neq \alpha'$. The action and observation spaces remain as they are in our original problem. The transition-observation distribution, given in Eqn. (3), is now time-invariant: we do not need to look forward in the search tree to compute the Q-values since the $\alpha$-vectors are available in the state space. Hence, this POMDP is well-defined.
    
    The dynamics of the POMDP are static and identical to the dynamics of our problem at time $t$. Thus, the backup operator $H_t$ is the backup operator for this POMDP and also for this POMDP's corresponding belief MDP. Therefore, the backup operator $H_t$ must be a contraction mapping. 
\end{proof}
\vskip 0.2in
\textbf{Theorem 4.} With suitable exploration, the value function constructed by our adapted POMCP algorithm converges in probability to the optimal value function, $V(h) \rightarrow V^*(h)$. As the number of visits $N(h)$ approaches infinity, the bias of the value function $\mathbb{E}[V(h) - V^*(h)]$ is $O(log(N(h))/N(h))$.
\begin{proof}
We will show that with enough exploration, in the limit of infinite samples, we have a well defined POMDP. The result then follows from Theorem 1 in \cite{silver2010monte}.

The human action nodes in the search tree maintain an array of values, which store the values of picking that action for different $\theta$. At any point in the search tree, the human actions (like the robot actions) are selected by picking the one that has the maximum augmented value (current estimated value plus exploration bonus). So, in the limit of infinite samples, each human action node is visited infinitely many times and the value estimates at the nodes converge to the true Q values. Having the correct human Q values gives us a POMDP, with well defined transition-observation dynamics. The result then follows from applying the analysis given in Theorem 1 of \cite{silver2010monte} to this POMDP.
\end{proof}

\newpage
\section{Pseudocode}
The pseudocode for adapted PBVI is presented below. The algorithm follows a similar structure to the standard PBVI algorithm \cite{pineau2003point}. The main difference between our adapted algorithm and the standard PBVI algorithm is that ours uses the modified Bellman update instead of the standard one. See lines 15-16.

The pseudocode for adapted POMCP is also presented below, similar in style to that presented in \cite{silver2010monte}. The key difference is that we maintain a live estimate of the sampled Q-values for $\Hu$ at each node. We maintain arrays which store the estimated values of taking each human action for different $\theta$. The optimal human action is selected with regard to these estimates. Like the robot actions, the human actions are selected while balancing exploration and exploitation. Rollouts use random action selection.
\begin{algorithm}[t!]
\caption{Adapted PBVI for CIRL Games}\label{pbvi}
\begin{algorithmic}[1]

\FUNCTION{PBVI ($b_0, T$)}
    \STATE $B \gets \{b_0\}$
    \STATE $V \gets$ Set of $\alpha$-vectors belonging to trivial plans
    \REPEAT
        \FOR{$t \in \{T-1, T-2, \ldots, 1, 0\}$}
            \STATE $V \gets$ Backup($B, V)$
        \ENDFOR
        \STATE $B \gets$ Expand($B, V$)
    \UNTIL{$\max_{\alpha\in V}\alpha\cdot b_0 \geq V_{target}$}
    \STATE \textbf{Return} $V$
\ENDFUNCTION
\\
~\\
\FUNCTION{Backup ($B, V'$)}
    \STATE $V \gets \{\}$
    \FOR{$a^R \in \AR$}
        \FOR{$\alpha_i'\in V'$}
            \FOR{$s \in S$}
                \STATE $Q_{H}(s, a^H) = \sum_{s'} T(s, a^H, a^R, s') \cdot$ \STATE $\ \ \ \ \alpha_{v(a^H)}(s')$
            \ENDFOR
            \STATE $\Gamma^{a^R} \gets \alpha_i(s) = r(s) + \gamma\cdot$
            \STATE $\ \ \ \sum_{a^H}\pi_{H}(a^H\mid Q_{H}(s, a^H))\cdot $
            \STATE $\ \ \ \ \sum_{s'} P(s', a^H\mid s, a^R, \alpha_i') \cdot \alpha_i(s')$
        \ENDFOR
    \ENDFOR
    \FOR{$b \in B$}
        \STATE $V_b \gets \{\}$
        \FOR{$a^R \in \AR$}
            \STATE $V_b\gets argmax_{\alpha\in\Gamma^{a^R}} \alpha\cdot b$
        \ENDFOR
        \STATE $V \gets argmax_{\alpha\in \Gamma_b} \alpha\cdot b$
    \ENDFOR
    \STATE \textbf{Return} $V$
\ENDFUNCTION
\\
~\\
\FUNCTION{Expand ($B', V'$)}
    \STATE $B \gets B'$
    \FOR{$b, \alpha \in B', V'$}
        \STATE $B_b \gets \{\}$
        \FOR{$a^R \in A^R$}
            \STATE $s \sim b(s)$
            \STATE $a^H \sim P(a^H \mid s,a^R, \alpha)$
            \STATE $b'(s') = \eta \sum_{s} P(s', a^H\mid s, a^R, \alpha) b(s)$ \STATE \ \ \ \ where $\eta$ is the normalizing constant
            \STATE $B_b \gets B_b \cup b'$
        \ENDFOR
    \STATE $B \gets B \cup argmax ||b-b'||_1, \forall b \in B_b, b' \in B'$
    \ENDFOR
    \STATE \textbf{Return} $B$
\ENDFUNCTION

\end{algorithmic}
\end{algorithm}
\newpage

\begin{algorithm}[t!]
\caption{Adapted POMCP for CIRL games}\label{adapted-pomcp}
\begin{algorithmic}[1]

\FUNCTION{Search ($h$)}
    \REPEAT
        \IF{$h = empty$}
            \STATE{$s \sim I$}
        \ELSE
            \STATE{$s \sim B(h)$}
        \ENDIF
        \STATE $\textsc{Simulate}(s,h,0)$
    \UNTIL{\textsc{Timeout()}}
    \STATE \textbf{Return} $argmax_{a^R}V(ha^R)$
\ENDFUNCTION
\\
~\\
\FUNCTION{Rollout ($s, h, depth$)}
    \IF{$\gamma^{depth} < \epsilon$}
        \STATE \textbf{Return} 0
    \ENDIF
    \STATE{$a^R, a^H \sim Uniform(A^R), Uniform(A^H)$}
    \STATE{$s' \sim T(s, a^R, a^H)$}
    \STATE{\textbf{Return} $r(s) + \gamma \cdot \textsc{Rollout}(s', ha^Ra^H, depth + 1)$}
\ENDFUNCTION
\\
~\\
\FUNCTION{Simulate ($s, h, depth$)}
    \IF{$\gamma^{depth} < \epsilon$}
        \STATE \textbf{Return} 0
    \ENDIF
    \IF{$h \notin T$}
        \STATE \textbf{Return} \textsc{Rollout}$(s,h,depth)$
    \ENDIF
    \STATE $a^R \gets argmax_{a^R}V(ha^R) + c\sqrt{\frac{\log N(h)}{N(ha^R)}}$
    \STATE $\theta \gets s_{\theta}$
    \STATE{$a^H \gets \textsc{SampleHumanAction}(\theta, h, a^R)$}
    \STATE{$s' \sim T(s, a^R, a^H)$}
    \STATE{$R \gets r(s) + \gamma \cdot \textsc{Simulate}(s', ha^Ra^H, depth + 1)$}
    \STATE $B(h) \gets B(h) \cup \{s\}$
    \STATE $N(h) \gets N(h) + 1$
    \STATE $N(ha^R) \gets N(ha^R) + 1$
    \STATE $N(ha^Ra^H) \gets N(ha^Ra^H) + 1$
    \STATE $V(ha^R) \gets V(ha^R) + \frac{R-V(ha^R)}{N(ha^R)}$
    \STATE $V_{\theta}(ha^Ra^H) \gets V_{\theta}(ha^Ra^H) + \frac{R-V_{\theta}(ha^Ra^H)}{N_{\theta}(ha^Ra^H)}$
    \STATE \textbf{Return} $R$
\ENDFUNCTION
\\
~\\
\FUNCTION{SampleHumanAction ($\theta, h, a^R$)}
    \STATE $a^H \sim \pi_H\Big (a^H\mid V_{\theta}(ha^Ra^H) + c\sqrt{\frac{\log N_{\theta}(ha^R)}{N_{\theta}(ha^Ra^H)}}\Big )$
    \STATE \textbf{Return} $a^H$
\ENDFUNCTION

\end{algorithmic}
\end{algorithm}

\clearpage
\section{Additional Experiments Omitted in Section 5}
\subsection{PBVI Experiment on Cooking Domain}
\begin{figure}[ht]
\begin{center}
\centerline{\includegraphics[width=\columnwidth]{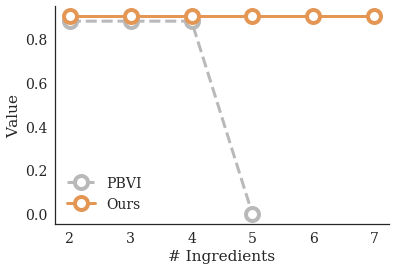}}
\caption{Value attained by PBVI and our approximate algorithm for various numbers of ingredients. (For the first 3 data points, the values attained by both methods were the same -- we jittered the plot slightly for visibility).}
\label{pbvi_ing}
\end{center}
\end{figure}

In our second of the three experiments conducted in Section 5, we compared the values attained by PBVI and our adaptation, with one hour of computation time. We first fixed the number of recipes and varied the number of ingredients. The results of this experiment are presented in \figref{pbvi_ing}. We found that both these algorithms, but especially our adapted algorithm, scaled much better with the number of ingredients than their exact VI counterparts. For simpler games, with 3 and 4 ingredients, both algorithms attained the maximal value of 0.9025. However, with 5 ingredients, PBVI found a value of 0 in an hour. In contrast, our algorithm easily solved the game with 5, 6, and 7 ingredients, attaining the maximal value of 0.9025. We next fixed the number of ingredients and varied the number of recipes. Again, our adapted algorithm outperformed PBVI. For example, with 4 recipes, our adapted method attains a value of 0.67875 while the standard PBVI method attains a value of 0.45125. 

These results suggest that our modified Bellman update allows PBVI to scale to larger CIRL games, especially in terms of the size of $\Hu$'s and $\R$'s action space. This offers further support to our hypothesis.
\newpage
\subsection{Experiment on \textit{RockSample} Domain}

\subsubsection{Domain}

The second benchmark CIRL domain we present is an extension of the POMDP benchmark domain \textit{RockSample}, that models Mars-rover exploration \cite{smith2004heuristic}. Consider a collaborative task where a human $\Hu$ wants to take samples of some rocks from Mars, with the help of a rover $\R$ deployed on Mars. There are some number of hours during the day (working hours) when $\Hu$ can control $\R$ herself but for the rest of the day, $\R$ has to behave autonomously. Not all types of rocks are of equal scientific value; $\Hu$ knows the value of each of these rocks but $\R$ does not. (Once again, we assume that $\Hu$ cannot communicate these values to $\R$ directly.)

Formally, consider an instance of \textit{RockSample} on a $m\times m$ grid, with $n$ rocks, each of which belong to one of $k$ different types. The state space is a cross-product of the x- and y-coordinate of $\R$ with $n$ binary features $IsSampled_i = \{Yes, No\}$, which indicate which of the rocks have already been sampled. (Each rock can only be sampled once.) 

\textit{RockSample} is a turn-based game: $\R$ may first take $l_{\R}$ steps in any of the four cardinal direction (during those hours when it is running autonomously) after which $\Hu$ may similarly take $l_{\Hu}$ steps (during the remaining hours). Thus, the set of actions available to $\Hu$ is the set of all trajectories of length exactly $l_{\Hu}$ while that available to $\R$ is the set of all trajectories with length at most $l_{\R}$. ($\R$ may wait on any specific step if it requires more information while $\Hu$ may not wait since she has all the information required.)

The set of all reward parameters $\Theta$ is composed of a collection of $k$-dimensional vectors, where the $i^{th}$ entry represents the reward received for sampling rock $i$. Both agents receive the reward specified by the true reward parameter $\theta$ when they sample a rock and receive no reward for any other action. 

\subsubsection{Details of Experiment}

We repeat the experiment from Section 5.1 of the main paper on this new domain, with a $5 \times 5$ grid ($m=5$), $3$ types of rocks ($k=4$), and $4$ rocks total ($n=4$).

This domain is much more complex than the cooking domain. For example, note that for even the simplest iteration of this domain, with $l_{\Hu} = l_{\R} = 1$, $|\AH| = 4$ and $|\AR = 5|$ (since $\R$ can also choose to wait); if we raise $l_{\R}$ slightly to 2, we have $|\AR = 13|$. Hence, we only ran our experiments with POMCP, which scales the best of the three types of the algorithms.

We found similar results to that of Section 5.2 of the main paper. For any value of $l_{\R}$ beyond $1$, FV-POMCP and POMCP fail to solve the problem; the branching factor of the search tree they both construct is too large and thus, both methods deplete the 16GB of memory in our system almost immediately. Our method however manages to scale to larger values of $l_{\Hu}$ and $l_{\R}$ with relative ease; our method successfully computed the optimal policy for this domain with values of $l_{\Hu} = l_{\R} = 2$ within two hours of computation.

\section{Additional Figure from Section 6.2}

\begin{figure}[ht]
\begin{center}
\centerline{\includegraphics[width=\columnwidth]{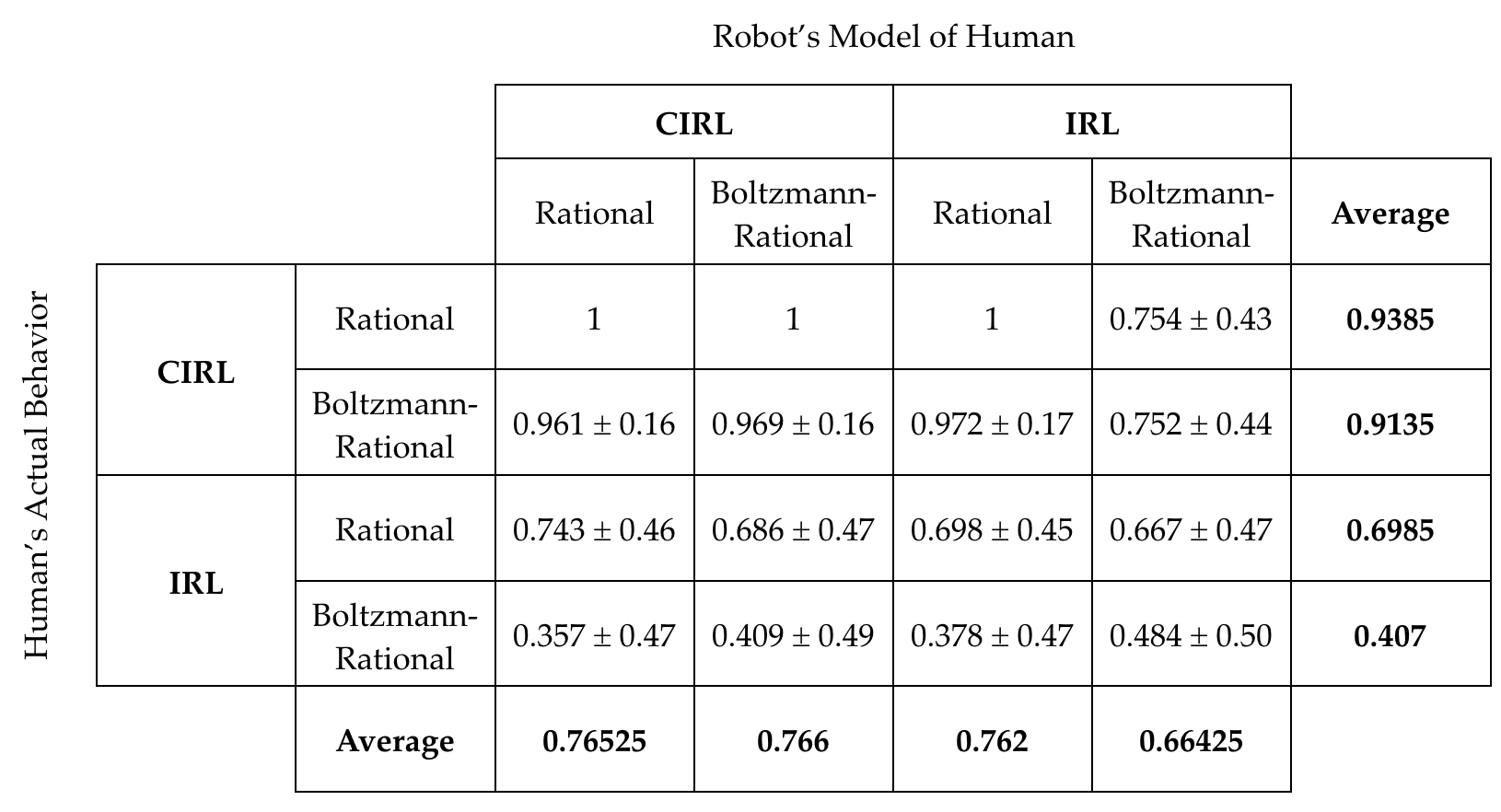}}
\caption{The proportion of times that $\Hu$ and $\R$ prepared the correct recipe on the cooking domain when $\R$ is trained with, and $\Hu$ actually behaves according to, a variety of different behaviors. They were significantly more successful at preparing the correct recipe when $\Hu$ behaved pedagogically according to CIRL. \vspace{-2em}}
\label{rational_comp}
\end{center}
\end{figure}

\figref{rational_comp} describes the results from the experiment in Section 6.2 of the main paper. In this experiment, we analyzed the performance of CIRL and IRL on our collaborative domain when the human does not behave rationally (and when the robot may or may not be aware of this fact).
\newpage
\section{Results of Follow-up Experiment to Section 6.2}

The manipulated variables and dependent measures are exactly as in Section 6.2, with the sole exception being that we considered 10 possible human policies instead of 2. The 10 policies were chosen from a $5\times 2$ factorial of the human's behavior and presence of bias. The 5 possible behaviors were rational, Boltzmann-rational with $\beta = 1$, Boltzmann-ration with $\beta = 5$, $\epsilon$-greedy with $\epsilon = 0.1$, $\epsilon$-greedy with $\epsilon = 0.01$. The 2 possible levels for presence of bias were "No Bias" and "Bias", where "Bias" denoted that $\Hu$ had a systematic preference for choosing the "Wait" action. (In our setting, $\Hu$ received a reward of 0.25 every time she chose the "Wait" action.)

\begin{figure*}[ht]
\begin{center}
\centerline{\includegraphics[width=\textwidth]{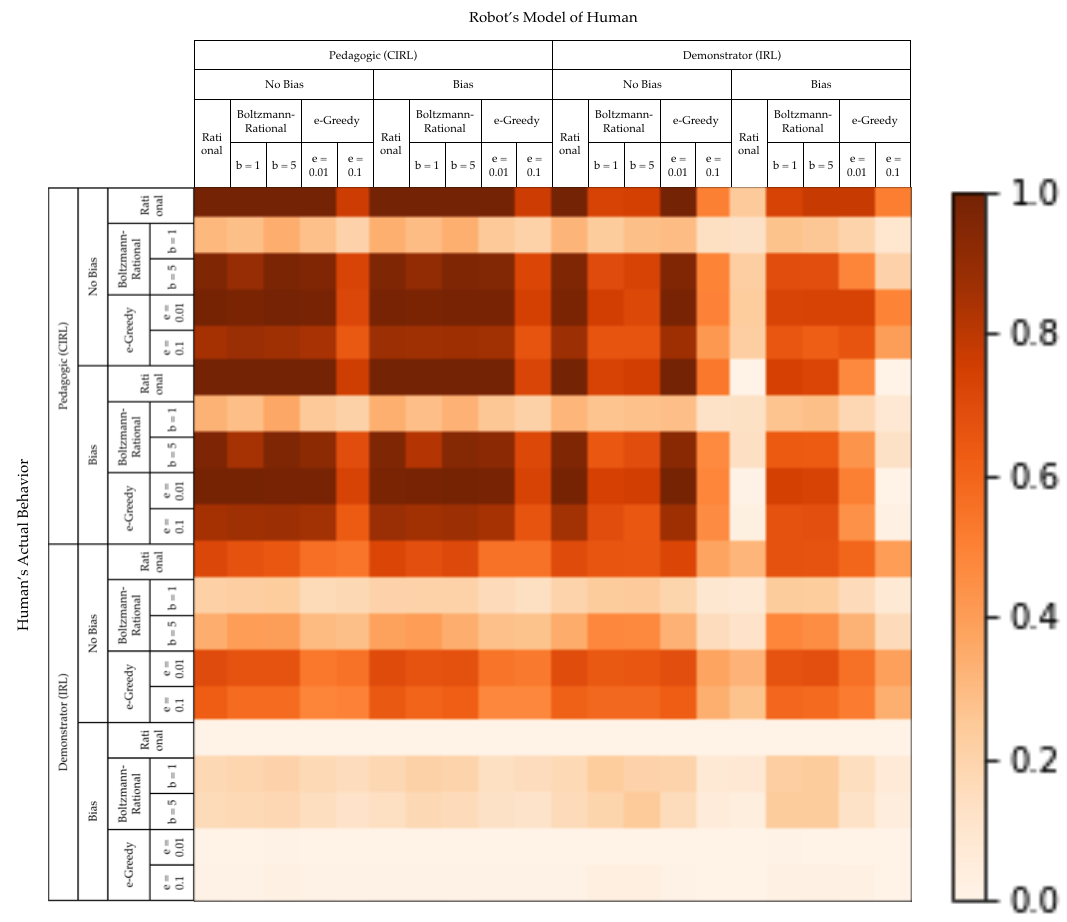}}
\caption{The proportion of times that $\Hu$ and $\R$ prepared the correct recipe on the cooking domain with 3 ingredients and 4 recipes when $\R$ is trained with, and $\Hu$ actually behaves according to, a variety of different behaviors. They were significantly more successful at preparing the correct recipe when $\Hu$ behaved pedagogically by following a policy specified by CIRL.}
\label{heat_map}
\end{center}
\end{figure*}

The results of our experiment are presented below in \figref{heat_map} as a heat map. Much like the simpler experiment in Section 7.2 of the main paper, we find that $\Hu$ and $\R$ are much more successful when $\Hu$ behaves pedagogically and that in the presence of pedagogic behavior, the team's performance is more robust to any sub-optimality from $\Hu$.

\clearpage

\section{Preliminary Human Subjects Study}

Previous work observed that CIRL has the potential to outperform standard IRL, and achieve value alignment by allowing a robot to exploit pedagogic behavior from humans \cite{hadfield2016cooperative}. However, CIRL was only empirically shown to improve upon the performance of IRL in theory or, as in our main paper, in simulation. This does not guarantee that we will observe a similar result in the real world, where the robot interacts with imperfect humans.

Our goal is to investigate whether the benefits of using CIRL over IRL in human-robot collaboration tasks carry over to practice. Here, we conduct a \textit{very preliminary} investigation into whether humans behave pedagogically in practice, and whether a robot trained with CIRL achieves value alignment more successfully than one trained with IRL.

\subsection{Hypotheses}

We anticipate that humans will objectively succeed at a collaborative task more frequently when collaborating with a CIRL robot as opposed to an IRL robot, especially when the task is complex. We also expect that humans will subjectively prefer to work with a CIRL robot instead of an IRL robot.

\noindent \textbf{H1 - Objective Performance.} \textit{The type of algorithm used will positively affect the collaboration objectively across a range of problem difficulties, with CIRL being better than IRL.}

\noindent \textbf{H2 - Objective Performance in Complex Problems.} \textit{On more complex problems, the type of algorithm used will positively affect the collaboration objectively, with CIRL being better than IRL.}

\noindent \textbf{H3 - Perceptions of the Collaboration.} \textit{The type of algorithm used will positively affect the participants' perception of the collaboration, with CIRL being better than IRL.}

\begin{figure}[t!]
\begin{center}
\centerline{\includegraphics[width=\columnwidth]{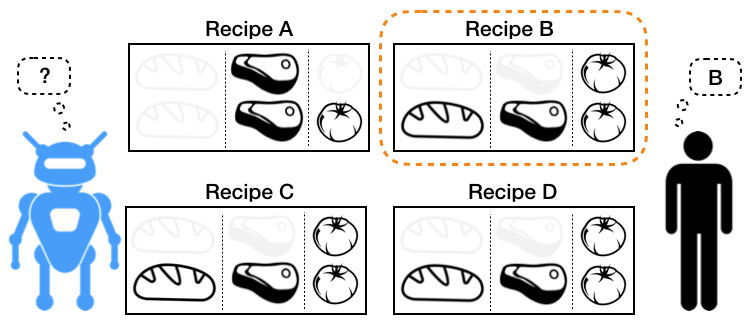}}
\caption{We conduct a \textit{very preliminary} investigation into whether humans do indeed behave pedagogically in practice and whether, as a result, CIRL is more effective than IRL for practically achieving value alignment. Participants collaborated with two robots, one trained with CIRL and another with IRL, to prepare a specified recipe, selected from a larger set. Both the participant and robot were allowed to make a single ingredient at each step but were only given two steps to complete the recipe; so, the human could not succeed without the robot's help. The robot did not know which recipe the participant was instructed to prepare. Participants therefore had to simultaneously teach the robot about their preferred recipe and make progress toward successfully preparing the recipe.\vspace{-2.5em}}
\label{front_fig}
\end{center}
\end{figure} 

\subsection{Experimental Design}

To explore the effect of the type of robot on human-robot collaboration, we conducted a counterbalanced within subjects study.

\subsubsection{Experimental Domain}

Participants collaborated with a virtual robot on the cooking task illustrated briefly in Figure \ref{front_fig} and described extensively in section 5.1 of the main paper. For this experiment, we kept the number of ingredients in the domain fixed at 3, and the length of the horizon fixed at 2.

The robot moved first in this domain. The human was allowed to observe the robot's move before selecting her own move.

\subsubsection{Manipulated Variables}

We manipulated two variables in our experiment. The first was the \textit{type of robot} used; the two levels were CIRL and IRL. We henceforth refer to a robot trained with CIRL as a CIRL-robot and similarly, to a robot trained with IRL as an IRL-robot.

We additionally wanted to investigate how both robots behaved across a variety of problem difficulties. We suspected that both robots would behave similarly on simple problems due to the straightforward nature of the tasks but that they would behave differently on more complex tasks where they could achieve the goal in a variety of ways. Hence, we additionally manipulated the \textit{number of recipes} used in the task from 2 to 5.

We initially attempted to also vary the length of the horizon on the collaborative tasks. However, we could only solve the longer horizon methods with POMCP, an approximate solver. Hence, there was no guarantee that the solution computed for these problems would be optimal or would be of similar quality across various runs. To avoid confounding the results, we chose to not vary the length of the horizon, and kept it fixed at 2.

We ran a full (2 by 4) factorial experiment with these two manipulated variables, leading to a total of 8 conditions.

\subsubsection{Procedure}

Participants entered the lab and were administered a pre-study questionnaire. Next, the experimenter explained the collaborative task and informed participants that they would be working with two different robots during the course of the experiment. The experiment was administered virtually -- the participants did not interact with physical robots.

They performed the task four times (one for each possible number of recipes) with one robot chosen at random, and then were administered a questionnaire and asked to describe the robot they had just worked with. They then performed the task four more times with the other robot and were similarly administered a questionnaire. They were finally administered a post-study questionnaire.

\subsubsection{Participant Assignment Method}

A total of 12 participants (10 males, 2 females, aged 18-25) were recruited from the local community. Ten of the participants reported having a technical background.

The experiment used a within-subjects design because it enables participants to compare the two robots. They were informed that one of the robots was a "student" robot that expected to be taught, and that the other was an "observer" robot that did not expect to be taught but would learn by watching the human perform the task as best as they could. They were made aware of which robot was the "student" and which was the "observer", so that they may behave accordingly and maximize their chance of succeeding at the task.

The order of the robot was counterbalanced to control for order effects. The recipe that participants were instructed to prepare in each condition was randomly chosen from the set of possible recipes to eliminate any systematic or familiarity bias.

\begin{table}[]
    \centering
    \begin{tabular}{l}
        \hline
         \textbf{Fluency}\\
         1. The human-robot team worked fluently together.\\
         2. The robot contributed to the fluency of the team\\interaction.\\
         \hline
         \textbf{Robot Contribution [shortened]} \\
         1. I had to carry the weight to make the human-robot\\team better.\\
         2. The robot contributed equally to the team\\performance.\\
         3. The robot’s performance was an important\\contribution to the success of the team.\\
         \hline
         \textbf{Trust [shortened]}\\
         1. I trusted the robot to do the right thing at the right\\time.\\
         2. The robot was trustworthy.\\
         \hline
         \textbf{Capability}\\
         1. I am confident in the robot’s ability to help me.\\
         2. The robot is intelligent.\\
         \hline
         \textbf{Predictability [rephrased for clarity]}\\
         1. The robot’s ingredient selection matched what I\\would have expected.\\
         2. The robot’s ingredient selection was surprising.\\
         \hline
         \textbf{Forced-Choice Questions}\\
         1. Which robot was the easiest to work with?\\
         2. Which robot do you prefer?\\
         \hline
         \hline
    \end{tabular}
    \caption{Subjective Measures}
    \label{list_questions}
\end{table}

\subsubsection{Dependent Measures}

\begin{figure*}[t!]
\begin{center}
\centerline{\includegraphics[width=0.9\textwidth]{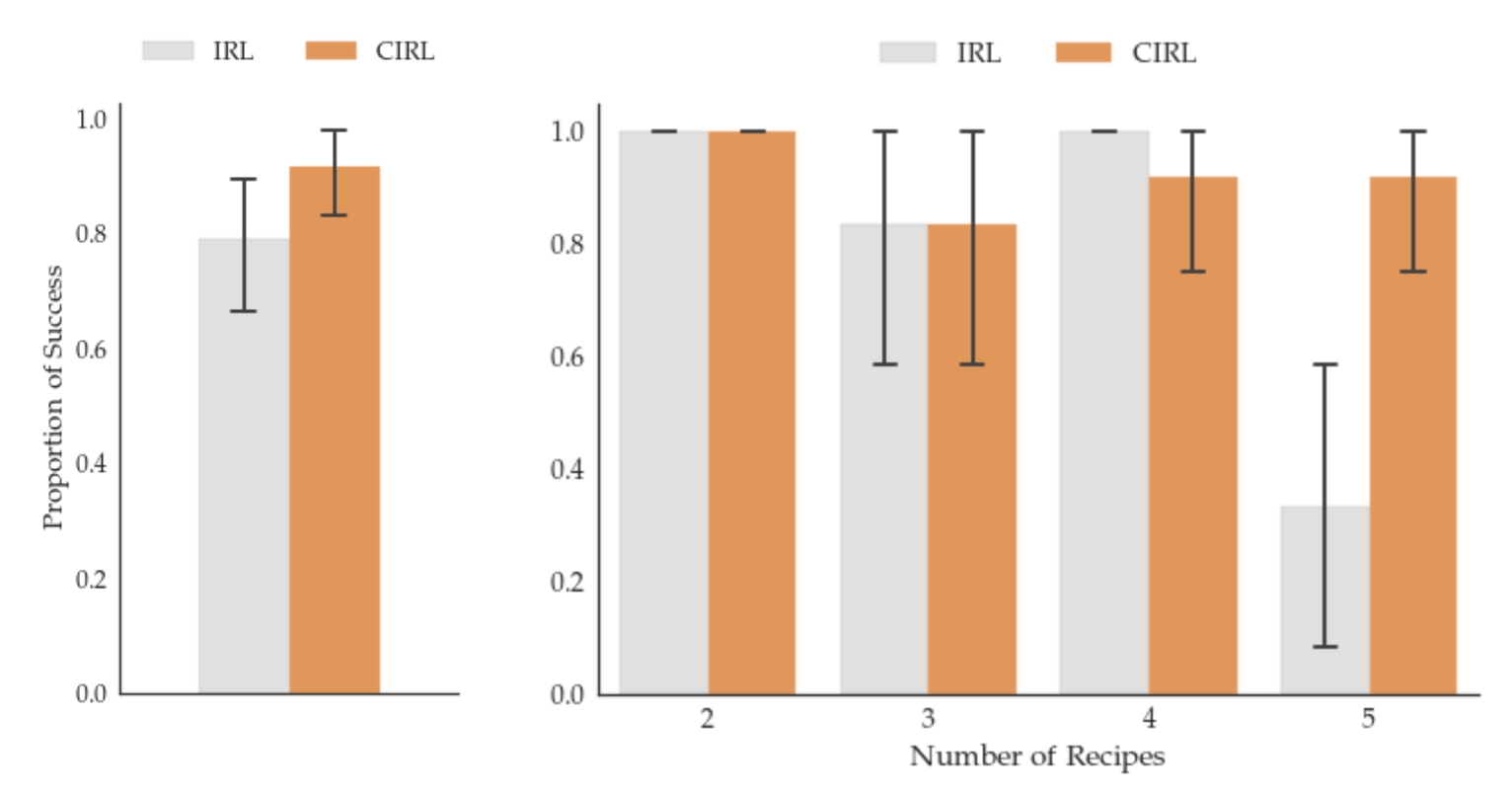}}
\caption{The proportion of all trials on which the human-robot team successfully prepared the correct recipe. (Left) Averaged across all number of recipes: the CIRL-robot only marginally outperforms the IRL-robot. (Right) For each number of recipes: the CIRL-robot only outperforms the IRL-robot for the most complex experiment (with 5 recipes). In all other experiments, both robots perform similarly.\vspace{-2em}}
\label{av_success}
\end{center}
\end{figure*} 

The measures capture the success of a collaboration in both objective and subjective ways, and are based on Hoffman's metrics for fluency in human-robot collaborations \cite{hoffman2013evaluating}.

The objective measure was \textit{success at preparing the desired recipe}. Participants were assigned a score of one when they succeeded and a score of zero when they failed.

Table 1 shows the six subjective scales that were used, together with a few forced-choice questions. We did not include the questions on Safety/Comfort since the participants did not interact with physical robots. The scales on Robot Contribution and Trust were shortened to avoid asking participants too many questions. The scale on Predictability was rephrased to more appropriately describe the setup of the experiment.

Additionally, participants answered forced-choice questions at the end, about which robot was easier to work with and which robot they preferred.

\subsection{Analysis}

\subsubsection{H1 - Objective Performance}

A repeated measures ANOVA on success at preparing the desired recipe showed that CIRL was only marginally better than IRL, when measured across all numbers of recipes ($F$(1,11) = 3.667, $p$ = 0.08). This offers some evidence in support of \textbf{H1}.

This is in line with the left plot in Figure \ref{av_success}, which show the results of the experiment. We see that the CIRL-robot outperforms the IRL-robot when averaged across all number of recipes but the improvement is marginal; the error bars for both algorithms have significant overlap.

\subsubsection{H2 - Objective Performance in Complex Problems}

A repeated measures ANOVA on success at preparing the desired recipe showed that there was a statistically significant interaction effect between the algorithm used and the number of recipes ($F$(1,11) = 16.18, $p$ = 0.002). A post-hoc analysis with Tukey HSD revealed that on complex problems with 5 recipes, CIRL significantly outperformed IRL, but on simple problem, there was no difference in performance between the two algorithms. This offers strong evidence for \textbf{H2}.

The right plot on Figure \ref{av_success} echoes these findings. On problems with 2, 3, or 4 recipes, the proportion of trials on which the CIRL-robot and IRL-robot prepared the correct recipe is very similar. However, on problems with 5 recipes, the CIRL-robot was much more successful that the IRL-robot in the collaboration task.

\subsubsection{H3 - Perceptions of the Collaboration}

\begin{table}[h]
    \centering
    \begin{tabular}{c|c|c|c}
        Scale &Cronbach's $\alpha$ &$F$(1,11) & p-value \\
        \hline
        \hline
        Fluency & 0.84 & 23.14 & \textbf{$<$0.001}\\
        Robot Contribution & 0.69& 17.73 & \textbf{0.001}\\
        Trust&0.89 & 16.95 & \textbf{0.002}\\
        Capability & 0.81 & 20.07 & \textbf{$<$0.001}\\
        Predictability & 0.86 & 5.189 & \textbf{0.04}\\
        Forced-Choice & 0.87 & 6.494 & \textbf{0.03}
    \end{tabular}
    \caption{Results of ANOVA on subjective metrics collected from a 7-point Likert-scale survey.}
    \label{results_sub}
\end{table}

\begin{figure*}[t!]
\begin{center}
\centerline{\includegraphics[width=0.9\textwidth]{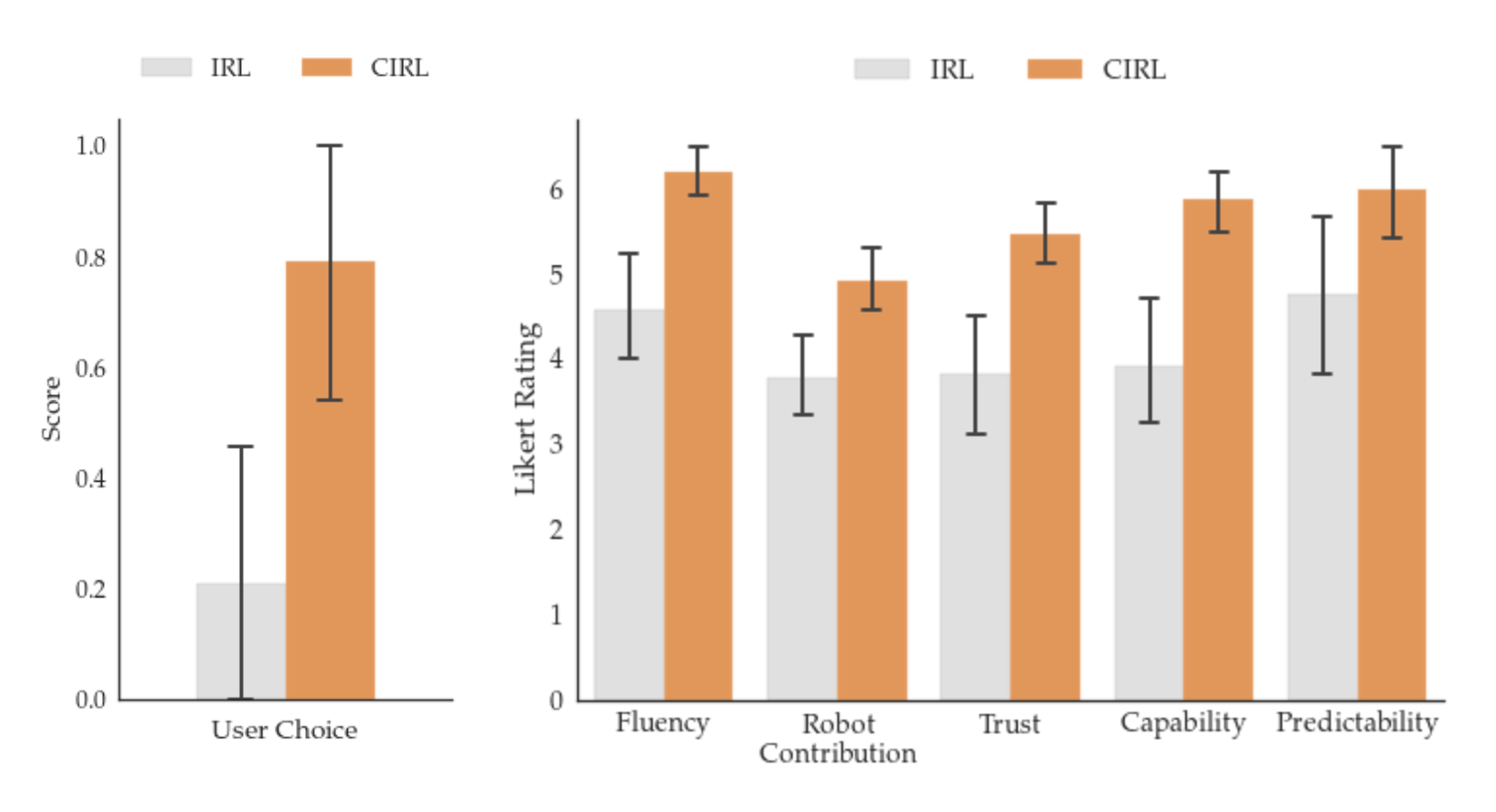}}
\caption{Findings for subjective measures. \vspace{-2em}}
\label{subjective_meas}
\end{center}
\end{figure*}

Table \ref{results_sub} shows the results of the experiment. The internal consistency of each scale is reported via Cronbach's $\alpha$. All scales except one had "good" consistency, the exception being robot contribution, whose consistency was "questionable" with a Cronbach $\alpha$ of 0.69. Scale items were combined into a score and analyzed with repeated-measures ANOVAs. Figure \ref{subjective_meas} plots the results.

The score produced by the overall forced-choice questions was significantly affected by the type of robot. The CIRL-robot had a significantly higher score than the IRL-robot ($p$ = 0.03); it was rated as being easier to work with by 9 of the 12 participants, and was preferred by 10 of the 12 participants. One participant rated the IRL robot as being easier to work with but preferred to work with the CIRL robot, remarking that he felt more "understood" by the CIRL robot.

All the Likert ratings showed a significant effect for type of robot as well; the CIRL-robot was rated significantly higher than the IRL-robot in \textit{every case} (with $p<$ 0.01 in all but one case -- predictability). The biggest difference between the two types of robot were in \textit{fluency} and \textit{capability} (both $p<$ 0.001). Several participants described the IRL robot as "not intelligent", with one remarking that she felt "the only reason we succeeded as much as we did was because some of the problems were so simple." 

These results offer strong evidence in favor of \textbf{H3.}

\subsection{Discussion}

We have empirically provided strong evidence that suggests that, in practice, CIRL is a more effective framework than IRL for value alignment. In our experiments, participants were objectively more successful at performing the specified human-robot collaboration task when working with a CIRL-robot than with an IRL-robot. Our results further suggest that CIRL leads to more fluent interaction between human and robot; our participants broadly preferred working with the CIRL robot than the IRL robot.   

Interestingly, when asked to describe their behavior, many participants described behaving similarly with both robots. One participant remarked that regardless of the robot she interacted with, she was "picking her ingredients to eliminate the wrong recipes as quickly as possible."

These remarks agree with the notion that humans tend to behave pedagogically when working with a learner in practice. It is then perhaps no surprise that CIRL significantly outperformed IRL -- by exploiting the pedagogic nature of humans, the CIRL-robot was able to infer more information more quickly than the IRL-robot was.

\subsubsection{Limitations and Future Work}

Due to the computational challenges of CIRL (outlined in the main paper), our experimental domain was still relatively straightforward. The short horizon nature of our task may made it easier for participants to behave pedagogically and questions remain as to whether people will behave similarly on more complex problems.

Additionally, the demographics of the participants of our survey were rather skewed toward males from technical backgrounds. It is entirely possible that people from technical backgrounds would be more informed about the behavior of the two robots and therefore able to more successfully collaborate with the robots than a non-technical person would.

In future work, we will explore how people behave in collaborative tasks over a long horizon, where their desire or ability to behave pedagogically may be impeded. Furthermore, we intend to deploy our algorithms on real robots and investigate how humans behave in collaborative tasks with actual robots as opposed to virtual ones on computer screens. To do so, we intend to develop a better online solution method for CIRL with stronger theoretical guarantees on the quality of the solution, thereby allowing us to solve larger problems in practice with real individuals.

\section*{Acknowledgements}

This work was supported in part by grants from the NSF NRI, and the Open Philantrophy Project.

\bibliography{example_paper}
\bibliographystyle{icml2018}

\end{document}